\pdfoutput=1

\documentclass[11pt]{article}

\usepackage[]{ACL2023}

\usepackage{pythonhighlight}
\usepackage{times}
\usepackage{latexsym}
\usepackage{inconsolata}
\usepackage{graphicx}
\usepackage{xspace}
\usepackage{hyperref}
\usepackage{enumitem}
\usepackage[normalem]{ulem}
\usepackage{inconsolata}
\usepackage{microtype}
\usepackage{inconsolata}
\usepackage{balance}
\usepackage[flushleft]{threeparttable}
\usepackage{enumitem}
\usepackage[linesnumbered,ruled,vlined]{algorithm2e}
\usepackage{algpseudocode}
\usepackage{booktabs}
\usepackage{subcaption}
\usepackage[nointegrals]{wasysym}
\usepackage{tikz}
\usepackage{tabu}
\usepackage{tabularx}
\usepackage{float}
\usepackage{listings}
\usepackage{color}
\usepackage{multirow}
\usepackage{framed}
\usepackage{pifont}
\usepackage{dsfont}
\usepackage{xcolor}
\usepackage{amsmath,amsthm}
\usepackage{amssymb}
\usepackage{graphicx}
\usepackage{textcomp}
\usepackage{tcolorbox}
\usepackage{soul}
\usepackage{makecell}
\usepackage[normalem]{ulem}
\usepackage{xspace}
\usepackage{arydshln}
\usepackage{setspace}
\usepackage{tcolorbox}
\usepackage{cleveref}

\usepackage[T1]{fontenc}

\usepackage[utf8]{inputenc}

\usepackage{microtype}

\usepackage{inconsolata}

\newcommand{\framework}[0]{\textsc{Sal}}

\title{Self-supervised Analogical Learning using Language Models}

\author{Ben Zhou\thanks{\ \ Work done while Ben Zhou was an intern at Amazon. All authors are/were affiliated with Amazon during this project.}\quad Sarthak Jain\quad Yi Zhang\quad Qiang Ning\quad \\ \textbf{Shuai Wang\quad Yassine Benajiba\quad Dan Roth}\\
Arizona State University\quad Amazon\quad University of Pennsylvania \\
\texttt{benzhou@asu.edu}\\
}

\begin{document}
\maketitle
\begin{abstract}
Large language models have been shown to suffer from reasoning inconsistency issues. That is, they fail more in situations unfamiliar to the training data, even though exact or very similar reasoning paths exist in more common cases that they can successfully solve. Such observations motivate us to propose methods that encourage models to understand the high-level and abstract reasoning processes during training instead of only the final answer. This way, models can transfer the exact solution to similar cases, regardless of their relevance to the pre-training data distribution. In this work, we propose \framework{}, a self-supervised analogical learning framework. \framework{} mimics the human analogy process and trains models to explicitly transfer high-quality symbolic solutions from cases that they know how to solve to other rare cases in which they tend to fail more. We show that the resulting models after \framework{} learning outperform base language models on a wide range of reasoning benchmarks, such as StrategyQA, GSM8K, and HotpotQA, by 2\% to 20\%. At the same time, we show that our model is more generalizable and controllable through analytical studies.
\end{abstract}

\section{Introduction}
\label{sec:intro}
\begin{figure}[t]
\begin{center}
    \includegraphics[scale=0.4]{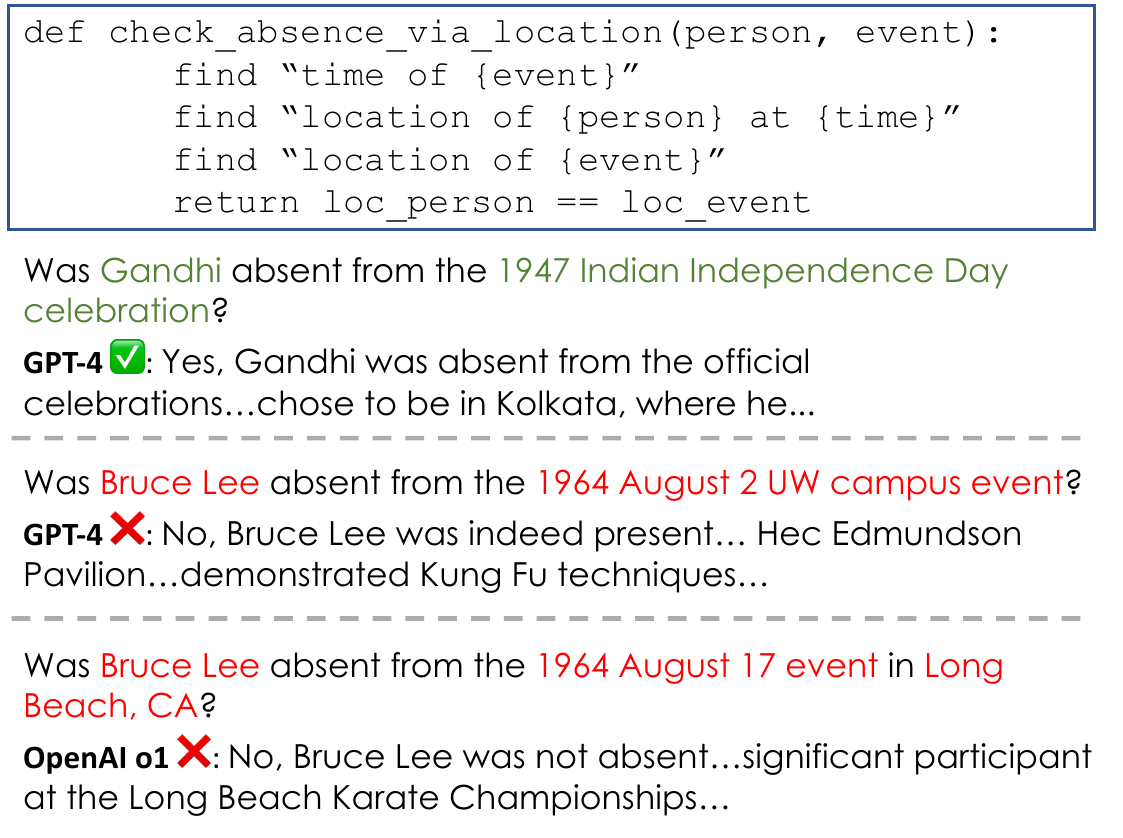}
    \caption{An illustration of models' (including o1) inconsistent reasoning processes for three questions that share the same problem-solving process (as outlined on top). Models memorize all relevant factual information (e.g., correctly answers where Bruce Lee was on those days). The correct answers are all ``yes,'' because Bruce Lee was in California on 1964.8.2, and he was in Seattle to get married on 1964.8.17.}
    \label{fig:example}
\end{center}
\end{figure}

While large language models (LLM) have achieved tremendous progress in reasoning benchmarks~\cite{Achiam2023GPT4TR}, they have been shown to have consistency issues~\cite{Li2023Deceiving, mirzadeh2024gsm} --- LLMs cannot produce consistent solutions on questions sharing identical or similar problem-solving processes. Previous studies~\cite{li2024famicom} have also shown that these models tend to perform well on cases involving familiar or common entities and words but fail on others. In Fig.~\ref{fig:example}, we demonstrate this issue with three questions that share an identical reasoning process (i.e., checking absence by checking location mismatch). The first question is a well-known fact that has been widely reported in written text, and models answer it correctly with the correct reason. The other two, on the other hand, involve less common entity combinations, and the models fail to answer them correctly, even though they possess all relevant knowledge by successfully pointing out where the person was at the time of the event. This consistency issue is not significantly improved even with newer models demonstrating much stronger reasoning performances: OpenAI o1\footnote{\url{https://openai.com/o1/}} also fails on the third question, albeit answering the previous two correctly. Such inconsistency prevents these models from deploying in mission-critical tasks (e.g., medical chatbots) requiring trustworthy and unbiased decision-making.

Following these observations, we are motivated to build models that can better understand the underlying high-level and abstract reasoning processes instead of only memorizing the final answers or relevant tokens during training time. This way, models can re-use such reasoning paths in all similar cases and achieve better consistency. 
To this end, we propose a data-driven analogical learning framework called \framework~(\textbf{S}elf-supervised \textbf{A}nalogical \textbf{L}earning). \framework{} extracts supervision signals with some inference processes using a base language model, and trains the same base language model with these signals, which forms a self-supervised learning scheme. The supervision signals we use are questions and their corresponding abstract and symbolic solutions, e.g., as shown on the top of Fig.~\ref{fig:example}, so that models can learn the high-level process that can bridge the inconsistency gap between various similar questions. 
We propose two novel supervision-extraction methods in \framework{}.
The first method called \textit{conceptualization}, starts with the original question and finds a group of similar questions that share \textbf{identical} high-level solutions~\cite{Zhou2023Symbolic} of the original question. With these questions, we collect symbolic solutions in the form of Python program from the questions that the base language model can successfully answer, and use these symbolic solutions as supervision targets. The second method, called \textit{simplification}, finds questions that share \textbf{sub-routines} of answering the original question, which can be used to eventually find the high-level solution of the original questions via composition. 

\framework{} collects supervision signals from both extraction methods and trains the base language model with the resulting supervision signals. At its essence, this training process enforces a soft but symbolic constraint over the questions that share similar or simpler problem-solving processes. We also design an optional factual retrieval step to help models better generate a single solution for complex questions where considering all possible related factors is infeasible. After training, \framework{} inference first generates a symbolic program and executes the program to find the final answer.
On a selection of complex reasoning datasets, including StrategyQA \cite{geva-etal-2021-aristotle}, ARC \cite{Clark2018ThinkYH}, CommonsenseQA \cite{talmor2018commonsenseqa}, GSM8K \cite{cobbe2021training}, and HotpotQA \cite{yang2018hotpotqa}, we show that \framework{} effectively improves base language models' performance with direct inference methods such as chain of thought and human-supervised baselines, by 2\% to 20\%. On top of the performance gains, \framework{} provides higher interpretability and controllability because of its programmatic inference scheme.\footnote{The code and data will be released upon publication.}

\begin{figure*}[t]
\begin{center}
    \includegraphics[scale=0.43]{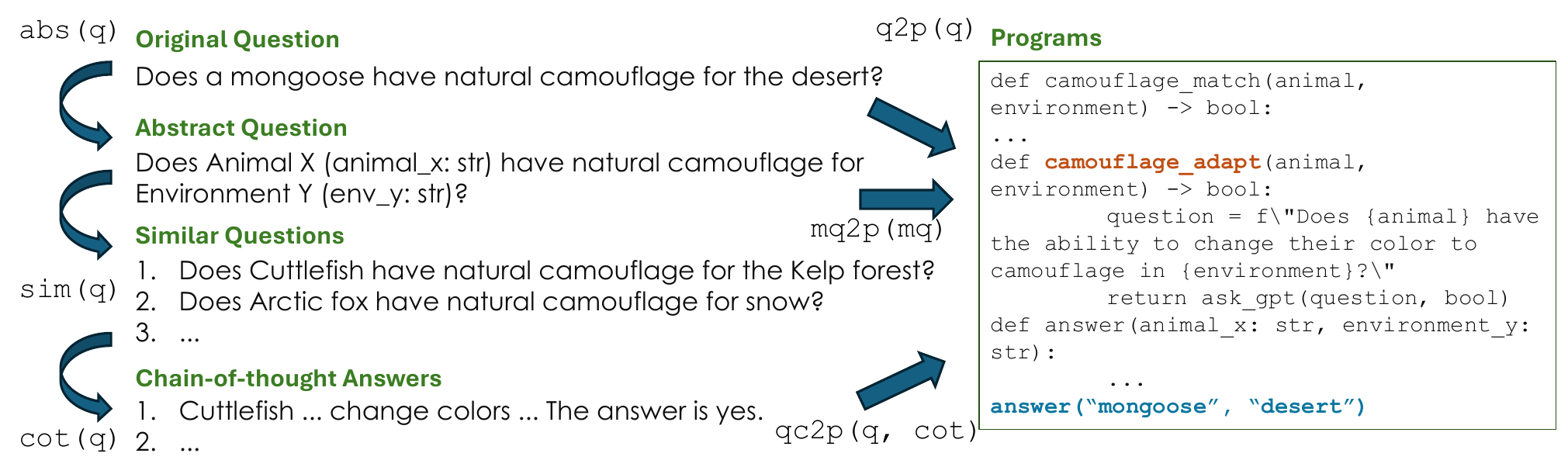}
    \caption{Overview of the conceptualization pipeline. Functions refer to sub-tasks and prompt templates.}
    \label{fig:binary}
\end{center}
\end{figure*}

\section{Related Work}

Our work builds on analytical works that discusses language model's inconsistency~\cite{tang2023large, Li2023Deceiving, li2024famicom, mirzadeh2024gsm}, and \framework{} is a learning scheme that mitigate such issues by encouraging consistency high-level solutions among relevant questions. \framework{} is related to previous works that use LLMs to generate programmatic solutions~\cite{chen2022program, gao2023pal} or other structures~\cite{feng2024bird}. Some of them also aims to improve model consistency~\cite{lyu2023faithful} via programs, but our work considers abstraction and high-level solutions, which can transfer better. Several works have also been using analogy to encourage generalization \cite{Hu2023InContextAR, Yasunaga2023LargeLM, Yuan2023ANALOGYKBUA, Zhou2023Symbolic}. Our work shares a similar motivation, but we are the first to propose automatic self-supervision methods and improve model performances in the training step. Our work is also related and motivated by other works that use self-supervision or distillation signals~\cite{zhu2024pad, zelikman2024quiet}. Decomposition-based inference pipelines~\cite{khattab2022demonstrate, hao2023reasoning, yaoreact} also inspires our work, as our programmatical solutions resembles a decomposed inference process.


\section{Self-Supervision Extraction}
\label{sec:supervision}

This section introduces how we collect the self-supervision signals in \framework{} through the two extraction methods: conceptualization and simplification. In both settings, we assume an original question $q$, and a base language model $LM$. Both extraction pipelines use $LM$ to find a set of questions $q'_{0}$, $q'_{1}$...$q'_{N}$ that share identical or sub-processes of the underlying problem-solving process of $q$. From $q'_{i}s$, we find a symbolic solution that describes the high-level reasoning process of $q$ that also applies to $q'_{i}s$ to some extent. We assume and require \textbf{no knowledge} regarding the gold answer or process of $q$, which makes \framework{} self-supervised.

The symbolic solutions we acquire are in the form of Python programs. The overall program structure design follows \citet{Zhou2023Symbolic} and supports general-domain question-answering (QA), including factual and math questions. On top of \citet{Zhou2023Symbolic} that uses $\mathrm{ask\_llm}$ calls for knowledge retrieval and soft value comparisons, we add a step to identify the units for physical values. This avoids a knowledge retrieval query such as ``What is the density of water?'' without providing the expected unit. We detail this process in \S\ref{sec:implementations}. All programs will contain a function named ``$\mathrm{answer}()$'', which will return a binary yes/no value for general questions and a specific number for math questions.

\subsection{Overview of Conceptualization}
\label{sec:overview-conceptualization}
The conceptualization extraction pipeline finds $q'_{0}$, $q'_{1}$...$q'_{N}$ that all share the same reasoning paths as $q$, but are much easier for $LM$ to confidently answer. 
Specifically, we first conceptualizes $q$ into a conceptualized version $q_{abs}$, as shown in Fig.~\ref{fig:binary}. Then, we generate a list of similar questions based on $q_{abs}$. With these questions, we select a set of high-confidence questions that language models tend to answer more accurately and use them as $q'_{i}s$. These components so far are primarily based on \citet{Zhou2023Symbolic}. Then, we propose a novel symbolic solution-generation process that generates high-quality programs from $q'_{i}s$ and their silver chain-of-thought outputs. We view these high-quality programs as supervision targets to $q$ and $q_{abs}$ to form self-supervision signals.

The intuition behind this is supported by recent works on LLM biases~\cite{Li2023Deceiving, li2024famicom}, where it is shown that LLMs do much better on questions that contain familiar entity and token combinations. Motivated by such observations, the conceptualization pipeline automatically creates easier-to-answer questions with identical or very similar solutions to $q$, acquiring high-quality symbolic solutions from these similar questions.

\begin{figure*}[t]
\begin{center}
    \includegraphics[scale=0.44]{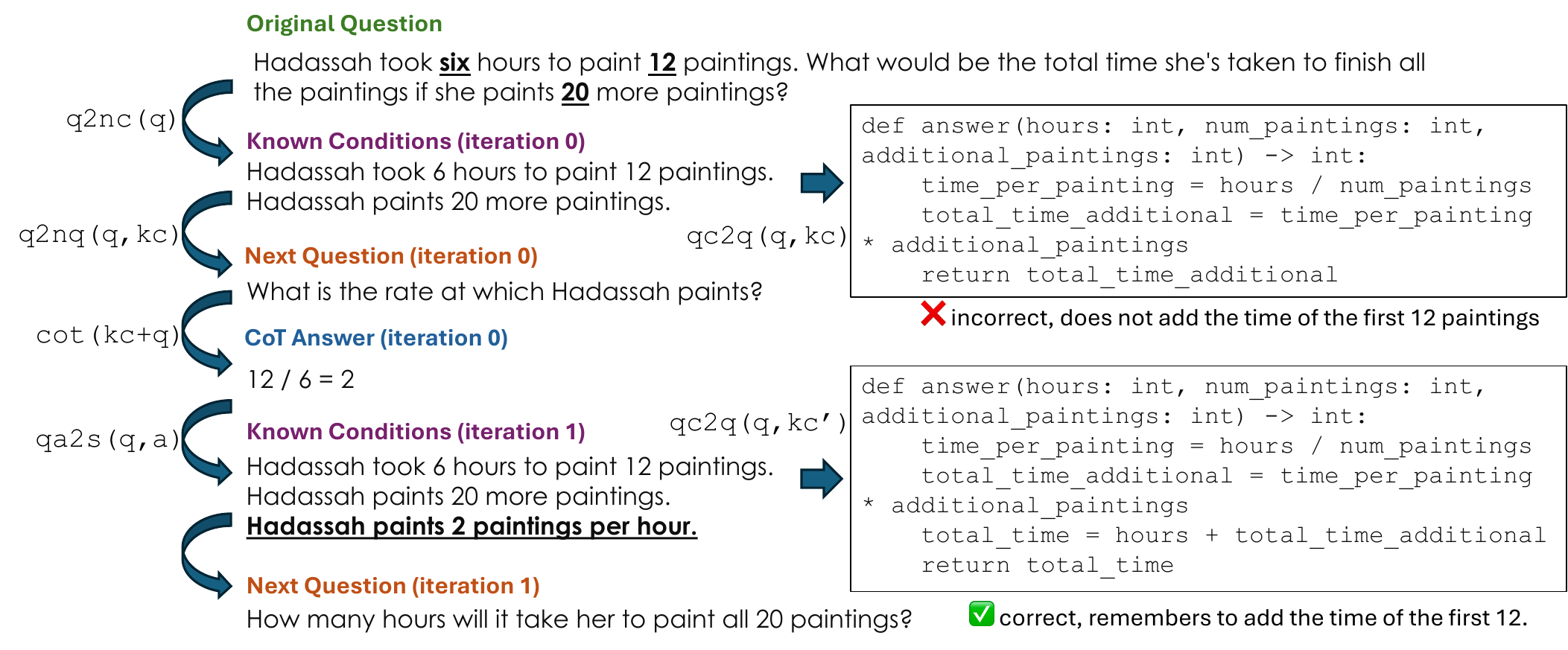}
    \caption{An example math question and overview of how our simplification process can acquire more correct incidental programs.}
    \label{fig:math}
\end{center}
\end{figure*}

\subsection{Overview of Simplification}
\label{sec:overview-simplification}
The simplification extraction targets math questions. We formulate solving a question $q$ as a step-wise decomposition process that iteratively finds the next single-hop question that is easier to answer and contributes to the final solution. We solve the next question and add the conclusion to a set of ``known conditions'' until the decomposition finds no new questions. These ``next questions'' are the $q'_{i}$s. Fig.~\ref{fig:math} provides an example of \textit{simplification}. The initial set of conditions are ``Hadassah took 6 hours to paint 12 paintings'' and ``Hadassah paints 20 more paintings.'' 
A valid next question would be ``What is the rate at which Hadassah paints?'' 

Our simplification method shares similarities in intuition with decomposition-based inference pipelines~\cite{hao2023reasoning}. Yet, we focus on acquiring a better global solution by reducing cognitive overload~\cite{xu2023cognitive}, which suggests that models perform better when they have fewer tasks simultaneously.
We observe similar findings where models produce better programs when fewer computations are needed for the current question.
For example, the most common mistake when solving the question in Fig.~\ref{fig:math} is that programs fail to add the 6 hours Hadassah took to paint the first 12 paintings, as the question asks for a total time. 
This is because the model has to deal with other sub-processes (i.e., computing the number of paintings per hour), reducing its capacity to think comprehensively. When provided with the results of part of the computations, models tend to think more comprehensively and generate better programs.

\subsection{Base Multi-function Model}
\label{sec:supervision:base-model}
As overviewed in \S\ref{sec:overview-conceptualization} and \S\ref{sec:overview-simplification}, we need multiple components (i.e., outputs from $LM$) in both extraction pipelines. We annotate a small set of domain-agnostic seed supervision signals for individual components to achieve this. This is because smaller models cannot sufficiently accomplish the specific sub-tasks we need even after regular instruction tuning. Larger language models can work with few-shot prompting and waive the need for such seed supervision. With such seed supervision signals, equivalent to instruction-tuning data, we train $LM$ to acquire an instruction-tuned base language model that can handle all sub-tasks needed in the pipelines. In this section, we introduce the individual components we need for both conceptualization and simplification extraction pipelines, as well as how we supervise $LM$ to achieve these components. Instruction templates are listed in \S\ref{sec:prompts}.


\paragraph{Chain of thought.} We collect seed supervision signals of questions to their chain-of-thought (CoT) answers, with a prompt template of $\mathrm{cot}(q)$, as shown in Fig.~\ref{fig:binary}. We borrow the CoT results of 228 binary questions from the StratgyQA development set from \citet{Zhou2023Symbolic} and acquire CoT outputs of an additional 265 math questions from the GSM8K training set using Mixtral-8x7B-Instruct.\footnote{\url{https://huggingface.co/mistralai/Mixtral-8x7B-Instruct-v0.1}}

\paragraph{Questions to abstract questions.} We collect seed supervision data of questions to their abstract or conceptual questions. We use the data from \citet{Zhou2023Symbolic} on the same 228 StrategyQA questions.
For the 265 math questions, we use Mixtral-8x7B-Instruct and few-shot prompting to generate similar abstractions. This process of generating abstractions is denoted as $\mathrm{abs}(q)$ in Fig.~\ref{fig:binary}.

\paragraph{Abstract questions to similar questions.} We use the data from \citet{Zhou2023Symbolic} to form seed supervision data of abstract questions to concrete similar questions, as $\mathrm{sim}(q)$ shows in Fig.~\ref{fig:binary}. 

\paragraph{Questions to programs.} We collect gold programs that can successfully solve the seed questions. We borrow the programs from \citet{Zhou2023Symbolic} for the 228 StrategyQA questions. We similarly use Mixtral-8x7B-Instruct to find candidate programs that execute the correct answer for math questions. With such programs, we can formulate two sub-tasks and their seed supervision, namely original question to program ($\mathrm{q2p}(q)$ as in Fig.~\ref{fig:binary}) and abstract question to program ($\mathrm{aq2p}(aq)$). We remove the ``answer()'' call in the target program in the second setting because the abstract question does not contain the specific input parameters.

\paragraph{Questions and relevant knowledge to programs.} We want to train a model sensitive to relevant knowledge and reference answers so that the model can generate precise and faithful programs when chain-of-thought answers or retrieval results are provided. For example, as Fig.~\ref{fig:binary} shows, we want the generated program based on the CoT results related to ``animal changing colors to adapt to an environment'' to \textbf{explicitly} consider whether an animal can change colors. To do this, we formulate seed supervision data that uses the questions and their chain-of-thought answers as inputs and corresponding programs as outputs. This subtask is denoted as $\mathrm{qc2p}(q, cot)$).

\paragraph{Original question to initial known conditions.} We generate seed supervision instances from original math questions to their initial sets of known conditions by few-shot prompting Mixtral-8x7B-Instruct using the 265 seed math questions in \S\ref{sec:supervision}. This task is denoted as  $\mathrm{math\_q2kc}(q)$. 

\paragraph{Question to next questions.} 
This task generates the next question based on the original and current known conditions.
The next questions are designed to be the immediate next computation step, given the current known conditions, to solve the original question. Similarly, we acquire seed supervision instances by few-shot prompting Mixtral-8x7B-Instruct with the 265 seed math questions. This task and its prompt template are named $\mathrm{math\_q2nq}(q, c)$, where $q$ is the original question, and $c$ is the current known conditions.

\paragraph{Question and answer to statements.} A final sub-task we define is transforming a question and its answer into a statement. We collect seed supervision data similarly by prompting Mixtral-8x7B-Instruct. This prompt template is named $\mathrm{qa2s}(q, a)$.


\paragraph{Final Model.} We train the base language model using the seed supervision data mentioned above, so this unified model can handle all sub-tasks as described by identifying the prompt templates we define in the seed supervision data. This is similar to all instruction-tuning methods that combine multiple sub-tasks during fine-tuning~\cite{khashabi2020unifiedqa, weifinetuned}.

\subsection{Conceptualization Extraction}

With such a multi-function model, we can now automatically extract self-supervision signals using the conceptualization pipeline overviewed in \S\ref{sec:overview-conceptualization}. For each question $q$, we generate its abstract question $q_{abs}$ by querying the multi-function model with template $\mathrm{abs}(q)$. We then generate a list of $N$ concrete questions that fit the abstract question $q_{abs}$ by prompting the model with the similar question generation prompt $\mathrm{sim}(q_{abs})$. For each concrete similar question $q'_i$, we acquire their chain-of-thought answers by prompting $cot(q'_i)$ and random sample $K$ times. We keep only the questions where $X$ out of the $K$ CoTs agree with each other to select high-quality questions that are easy and confident enough for the model to answer. For each of the selected similar questions $q'_i$, we find one of the CoT results $cot'_i$ that infer to the majority-voted answer and generate a programmatic solution using $p_{i}=\mathrm{qc2p}(q'_i, cot'_i)$. 

\subsubsection{Selection Criteria and Parameters}
In this work, we set $N=20$, $K=10$, and $X=9$, meaning that we generate 20 similar questions per abstract question and keep those with at least a 9/10 agreement in their chain-of-thought predictions. After the inference process, we would have a set of programs generated from these high-confidence similar questions. Because we have multiple programs from this generation process, we are motivated to further select from them and use the ones with the highest quality possible as the self-supervision signals. We now describe a set of selection criteria we use for this purpose. All hyper-parameters are tuned with manual examinations over a small set of examples, and we expect minimum variances because of the large data sizes.

\paragraph{Accuracy-based criteria.} A natural selection process is to verify each programmatic solution against the high-confidence similar questions by trusting their corresponding silver labels. We execute each of the generated programs with the parameters corresponding to each selected similar question. Let the number of remaining similar questions be $M$; we set a high threshold of $max(3, M * 0.75)$ and select programs that execute to the correct silver labels\footnote{Here the silver labels refer to the CoT answers of each generated similar questions, as we do not use gold labels of the original questions in any way.} over such a threshold number of questions. We also employ a low threshold of $2$, which will be used later. 

\paragraph{Similarity-based criteria.} We employ a similarity-based metric using a paraphrase model from sentence transformers\footnote{\url{https://huggingface.co/sentence-transformers/paraphrase-MiniLM-L6-v2}} to encourage the model to generate programs that do not repeat the original question and perform actual decomposition and reasoning. We drop all programs containing a $\mathrm{ask\_llm}$ query with a paraphrase score higher than $0.95$.

\paragraph{Complexity-based criteria.} We also want the model to self-supervise on programs considering more abduction factors, which implies more query calls to $\mathrm{ask\_llm}$. As a result, we drop all programs that only make one query using this helper function.

\paragraph{Soundness-based criteria.} We also check for details in the generated program to ensure soundness. Specifically, we drop all programs that contain unused parameters in the $\mathrm{answer}()$ function call and all programs that produced an error message when executing on more than two similar questions.

\paragraph{} We formulate two kinds of self-supervision instances from the selected programs. The first is question-to-program, where the corresponding similar question is used as input, and we use the $\mathrm{q2p}()$ prompt template and the low accuracy threshold described above as long as the program successfully executes the correct silver answer on that specific similar question. The second is abstract-question-to-program, using the high accuracy threshold to ensure the target program is representative enough for all scenarios and the $\mathrm{aq2p}()$ input template.

\subsection{Simplification Extraction}




At inference time, for each math question $q$, we first generate the initial set of known conditions by prompting the base multi-function model with $\mathrm{math\_q2kc}(q)$. Then, we propose an iterative process with the following steps. 1) Generate the next question using $\mathrm{math\_q2nq}(q, c)$; 2) Answering the next question with $\mathrm{cot}(kc+nq)$, where we concatenate the current known conditions with the next question into a single query; 3) Adding the next question and its question to the set of known conditions by transforming them into a statement using $\mathrm{qa2s}(nq, a)$. 4) Generating solutions in our programmatic space using $\mathrm{qc2p}(q, kc)$, where $q$ is the original question, and $kc$ is the current known conditions. The iteration stops if one of the two criteria happens. A) The generated next question contains ``no more decomposition,'' which is part of the above seed training data. B) 9/10 of the generated programs from 4) execute the same answer, which implies high confidence in the predictions.

\paragraph{Sanity-check Experiments}

\begin{table}[t]
\centering
\small
\begin{tabular}{lccc}
\toprule
System & \#Instances & Accuracy & Collected Pct. \\
\cmidrule(lr){1-1}\cmidrule(lr){2-2}\cmidrule(lr){3-3}\cmidrule(lr){4-4}
Original & 344 & 91.3 & 25.3\%\\
Iteration 1 & 103 & 86.4 & 32.9\% \\
Iteration 2 & 92 & 90.2 & 39.7\% \\
Iteration 3 & 46 & 87.0 & 43.1\% \\
Iteration 4 & 15 & 80.0 & 44.2\% \\
Iteration 5 & 1 & 100.0 & 44.3\% \\
\bottomrule
\end{tabular}
\caption{Number of math questions that produced ten programs that execute to the same answer at each timestamp of the simplification iteration. Collected Pct. indicates the number of accumulated questions from which we can acquire high-confidence programs.}
\label{tab:math-sanity}
\end{table}

We conduct a sanity-check experiment to verify the effectiveness of our proposed simplification method for mathematical questions. We aim to acquire as many high-confidence programs as possible and evaluate the quality of these high-confidence programs. We run our proposed simplification pipeline on 2k math questions from GSM8K, among which 1,358 questions produced valid programs. Because we want to collect high-confidence programs, we count the number of math questions that can produce ten programs that lead to the same answer after execution. As Table~\ref{tab:math-sanity} shows, without our simplification pipeline, only on 25.3\% of the questions can the model generate ten programs that lead to the same answer, which we consider as a valid self-supervision instance. At the 5th iteration of our simplification process  (i.e., generating 5 next questions iteratively), we can collect high-confidence programs on 44.3\% of the math questions, with high accuracy on par with the original generation (i.e., generating programs directly from the original question). This suggests that our proposed method can effectively acquire more self-supervision signals without undermining their qualities.

\begin{table*}[ht]
\centering
\small
\begin{tabular}{lccccccc}
\toprule
System & StQA-seen & StQA & GSM-seen & GSM & ARC & CsQA & Hotpot \\
\cmidrule(lr){1-1}\cmidrule(lr){2-2}\cmidrule(lr){3-3}\cmidrule(lr){4-4}\cmidrule(lr){5-5}\cmidrule(lr){6-6}\cmidrule(lr){7-7}\cmidrule(lr){8-8}
CoT (few-shot) & 58.8 & 58.8 & 44.6 & 39.2 & 70.5 & 62.2 & 32.7 \\
Seed Only & 57.6 & 56.1 & 45.4 & 45.0 & 68.3 & 73.5 & 50.3 \\
\midrule
\multicolumn{8}{c}{\textbf{Ours}}\\
\midrule
\framework{} (concept. only) & 61.1 & 58.8 & 44.7 & 43.7 & 68.3 & 72.0 & 51.5 \\
\framework{} (all) & 61.0 & 59.7 & 49.4 & 50.8 & 71.0 & 74.7 & 52.4\\
\framework{} (all) + RAG & 63.7 & 62.9 & 49.4 & 50.8 & 73.7 & 77.0 & 55.1 \\
\bottomrule
\end{tabular}
\caption{System performances on a range of reasoning datasets. The base model is Mistral-7B. Few-shot CoT uses Mistral-7B-Instruct-v0.1.}
\label{tab:main}
\end{table*}

\begin{table}[ht]
\centering
\small
\begin{tabular}{lccc}
\toprule
System & StQA-seen & StQA & Avg. $\Delta$\\
\cmidrule(lr){1-1}\cmidrule(lr){2-2}\cmidrule(lr){3-3}\cmidrule(lr){4-4}
Seed Only & 45.8 & 45.8 & -11.1 \\
\framework{} & 58.0 & 56.4 & -3.2 \\
\bottomrule
\end{tabular}
\caption{System performances after removing all programs that internally query the language model with a question very similar to the original question.}
\label{tab:no-sim}
\end{table}

\begin{table*}[ht]
\centering
\small
\begin{tabular}{lccccc}
\toprule
System & StQA Dev. & StQA test & BoolQ & CsQA & Hotpot \\
\cmidrule(lr){1-1}\cmidrule(lr){2-2}\cmidrule(lr){3-3}\cmidrule(lr){4-4}\cmidrule(lr){5-5}\cmidrule(lr){6-6}
Mixtral 8x7B (few-shot) & 53.9 & 52.9 & 41.1 & 10.7 & 55.3 \\
Llama-70B (few-shot) & 56.8 & 59.7 & 58.9 & 53.4 & 65.6 \\
\cmidrule(lr){1-6}
Mistral7B + Seed Only & 60.1 & 44.3 & 40.1 & 59.0 & 50.0 \\
Mistral7B + \framework{} & 59.9 & 56.0 & 51.2 & 59.0 & 51.2 \\
\bottomrule
\end{tabular}
\caption{Conceptual Reasoning Performance. We use a slightly different program space, as described in \S\ref{sec:supervision}.}
\label{tab:conceptual}
\end{table*}

\section{Training with Analogical Supervision}
As mentioned in \S\ref{sec:intro}, \framework{} first collects self-supervision signals and then trains the model with them. We now describe the training process.

\subsection{Direct Training}
\label{sec:framework:direct}
The direct training scheme simply combines all the self-supervision instances and the seed training instances and trains a new model with a supervised fine-tuning (SFT) process. Our goal is to show that the models trained with our automatic self-supervision data can outperform two baselines: 1) chain-of-thought inference with the base language model's instruction-tuned version, 2) the base language model supervised with the seed supervision from \S\ref{sec:supervision:base-model} only.

\subsection{Retrieval-augmented Generation}
\label{sec:framework:retrieval}
The reasoning process sometimes requires some prior knowledge to reduce the size of the abduction. For example, consider the question, ``Can I visit Museum X 20 times for under 200 dollars?'' When generating programs, the model will almost always consider the ticket price for a single entry because it is the most common pricing structure. However, there is a complicated pricing structure for certain museums, such as the MET museum, including annual memberships, senior and student discounts, and residential passes. It is impractical for models to consider all possible pricing structures in the first pass. We build a retrieval-augmentation mechanism to allow models to consider external world facts before generating the solutions, reducing the search space and improving the feasibility of generating a single program to solve the question. To achieve this, we take all the $ask\_llm$ queries in the generated programs from the first pass and ask the base language model to answer them with a short paragraph. Then, we use $\mathrm{qc2p}(q, ans)$ to generate the second-pass programs, where $ans$ is the short answering paragraph. Whenever a voted answer from the first-pass programs is ``unknown,'' we use the retrieval-augmented programs instead.

\section{Experiments}
In this section, we conduct experiments to show the effectiveness of our proposed framework \framework{}.

\subsection{Self-Supervision Extraction}
The seed supervision data contains 4619 instances that have programs as outputs. To expand them with self-supervision, we apply the conceptualization pipeline on 2k questions from StrategyQA and 800 questions from HotpotQA. We apply simplification on 5k questions from GSM8K. All questions come from the training sets. Such processes produced 5865, 1000, and 4439 self-supervision training instances, respectively.

\subsection{Baselines, Datasets and Parameters}
Due to resource concerns and other constraints, we use Mistral-7B-v0.1 as our base language model. Both the training for the multi-function model in \S\ref{sec:supervision:base-model} and later experiments on evaluating the effectiveness of \framework{} supervision signals are done with the same model because of the self-supervision assumption. $\mathrm{ask\_llm}$ is handled by few-shot prompting to its instruction-tuned version (Mistral-7B-Instruct-v0.1). All fine-tuning processes use LoRA \cite{hu2021lora} (r=32) and a learning rate of $2e-4$.
We use $k=10$ as the number of candidate programs at inference time.

We evaluate five reasoning datasets: StrategyQA, GSM8K, ARC, CommonsenseQA, and HotpotQA. Because of the limitations of our program space (i.e., only support binary answers), we transform all questions to binary for multiple-choice datasets. Specifically, we design a few-shot prompt and use Mixtral-8x7B-Instruct to transform the question with its correct answer to a question with a correct answer of ``yes''. As this is a simple rephrasing task, Mixtral demonstrates 100\% accuracy in the 50 human-selected examples. For StrategyQA and GSM8K, we create a special evaluation split called ``seen'', which contains questions we use as raw questions when generating self-supervision signals. Models do not see or use the gold labels of these questions in any way. The overall evaluation data contains 500 StrategyQA (seen), 490 StrategyQA, 500 GSM (seen), 500 GSM, 112 ARC, 336 CommonsenseQA and 339 HotpotQA instances. 

We consider two main baselines: a few-shot chain-of-thought (\textbf{CoT}, self-consistency@10) of Mistral's instruction-tuned version, and the base model trained on seed supervision instances (\S\ref{sec:supervision:base-model}) only, named ``\textbf{Seed Only}.'' All models are trained to the same steps, which equals five epochs of the full training data (seed + self-supervision). We report averaged results from two random seeds for experiments that require SFT.

\subsection{Results}
Table~\ref{tab:main} shows our main experiment results of reasoning performances across the five reasoning datasets. We first observe that the seed only baseline performs relatively similarly to the CoT baseline, indicating that the seed supervision data mostly teach the model forms of the programmatic space without improving their reasoning capabilities by much. On the other hand, we see that the base LLM trained on self-supervision data from conceptualization extraction only (\framework{} (concept. only)) improves 3.5\%, 2.7\%, and 1.2\% on StrategyQA's seen, unseen, and HotpotQA, demonstrating the effectiveness of our proposed self-supervision signals. By adding the simplification self-supervision (\framework{} (all)), we improve 4\% and 5.8\% on GSM8K's seen and unseen sets, showing that our simplification pipeline can effectively extract self-supervision data on math-like questions. In addition, we observe better domain transferability as \framework{} (all) improves on ARC and CommonsenseQA by 2.7\% and 1.2\%, respectively. Model performance is further improved by adding the retrieval-augmented generation in \S\ref{sec:framework:retrieval}. The final model is consistently better than the CoT baseline (+6.8\% on StrategyQA, 5.8\% on GSM8K, and 4.8\% on HotpotQA). At the same time, our model is consistently better than the seed only baseline, demonstrating the overall effectiveness of our proposed self-supervision framework.

\section{Analysis}
To further demonstrate the capability of our self-supervision signals, we conduct an experiment that disregards all programs that contain a $\mathrm{ask\_llm}$ query that is at least 0.9 similar to the original question, using the same paraphrase detection model in \S\ref{sec:supervision}. As the results show in Table~\ref{tab:no-sim}, the seed baseline model drops significantly by 11\% on StrategyQA, meaning that a significant portion of its generated programs simply repeat the original question and delegate reasoning to the instruction-tuned model behind $\mathrm{ask\_llm}$. This is not as much a concern in our self-supervised model as it only drops 3\%. This suggests the self-supervised model is better at reasoning by conducting comprehensive abduction and decompositions.

We also consider the conceptual reasoning benchmark proposed in \citet{Zhou2023Symbolic}, which considers four datasets and 1597 test instances. In this conceptual setting, systems see an abstract version of the question and must make conceptual reasoning considering all possible corner cases, which we can achieve by querying our models with $\mathrm{aq2p}(aq)$. 
As Table~\ref{tab:conceptual} shows, the model trained with self-supervision signals improves 11.7\% on StrategyQA, 11\% on BoolQ, and 1.2\% on HotpotQA. At the same time, it outperforms the larger Mixtral-8x7B-Instruct model on several tasks, demonstrating the effectiveness of using automatic self-supervision signals. Such high performance on conceptual reasoning suggests that our proposed self-supervised model improves neutral and unbiased reasoning by establishing internal consistencies in the actual reasoning processes.

\section{Conclusion}
This work introduces \framework{}, self-supervision analogical learning that encourages LLM consistent reasoning in common and rare situations. We propose two methods to collect self-supervision signals: 1) conceptualization, a pipeline that finds programmatic solutions from similar questions that models can confidently answer and share identical reasoning processes; 2) simplification, which finds high-quality solutions to math questions by gradually decomposing a question and reducing models' logical overload. Experiments show that the self-supervision signals are high-quality and effective, which can be used to train models more attentive to the actual reasoning process instead of the semantic inductive biases. \framework{} leads to performance improvements on a wide-range of reasoning tasks.

\section*{Limitations}
Our work has several limitations in its current form.

\noindent
\textbf{Limited Data Sizes.} We only consider 3k questions for the conceptualization pipeline and 5k questions for the simplification pipeline. Because of the selection criteria, the resulting data size is only 2.5 times larger than the seed supervision. This limits our arguments on large-scale self-supervision. 

\noindent
\textbf{Binary Questions Only.} We only consider binary questions in the conceptualization pipeline because it is the most straightforward way to evaluate a program solution's accuracy on similar questions. While all multiple-choice questions can be easily transformed into binary questions, this limits the scalability to more free-form questions and needs to be addressed in future works.

\noindent
\textbf{Single Base Model.} Due to computational constraints, we only experiment with one base LLM (Mistral-7B), and more extensive experiments with other families of models may provide more insight into the capabilities of different models. Yet, we hypothesize that the general trend should be consistent with other base models.

\noindent
\textbf{Single Base Model.} Due to computational constraints, we only experiment with one base LLM (Mistral-7B), and more extensive experiments with other families of models may provide more insight into the capabilities of different models. Yet, we hypothesize that the general trend should be consistent with other base models.

\noindent
\textbf{Limited Ablation Studies.} Due to computational constraints, we do not conduct ablation studies on the hyper-parameters used in \S\ref{sec:supervision}, although these parameters can be further fine-tuned to achieve better performances. However, we believe that the hand-picked parameters we employ are good enough because of manual examinations, and we expect little changes in empirical results if these numbers are changed slightly because we extract a relatively large set of supervision instances.

\bibliography{anthology,custom}

\begin{thebibliography}{27}
\expandafter\ifx\csname natexlab\endcsname\relax\def\natexlab#1{#1}\fi

\bibitem[{Chen et~al.(2022)Chen, Ma, Wang, and Cohen}]{chen2022program}
Wenhu Chen, Xueguang Ma, Xinyi Wang, and William~W Cohen. 2022.
\newblock Program of thoughts prompting: Disentangling computation from reasoning for numerical reasoning tasks.
\newblock \emph{arXiv preprint arXiv:2211.12588}.

\bibitem[{Clark et~al.(2018)Clark, Cowhey, Etzioni, Khot, Sabharwal, Schoenick, and Tafjord}]{Clark2018ThinkYH}
Peter Clark, Isaac Cowhey, Oren Etzioni, Tushar Khot, Ashish Sabharwal, Carissa Schoenick, and Oyvind Tafjord. 2018.
\newblock \href {https://api.semanticscholar.org/CorpusID:3922816} {Think you have solved question answering? try arc, the ai2 reasoning challenge}.
\newblock \emph{ArXiv}, abs/1803.05457.

\bibitem[{Cobbe et~al.(2021)Cobbe, Kosaraju, Bavarian, Chen, Jun, Kaiser, Plappert, Tworek, Hilton, Nakano et~al.}]{cobbe2021training}
Karl Cobbe, Vineet Kosaraju, Mohammad Bavarian, Mark Chen, Heewoo Jun, Lukasz Kaiser, Matthias Plappert, Jerry Tworek, Jacob Hilton, Reiichiro Nakano, et~al. 2021.
\newblock Training verifiers to solve math word problems.
\newblock \emph{arXiv preprint arXiv:2110.14168}.

\bibitem[{Feng et~al.(2024)Feng, Zhou, Lin, and Roth}]{feng2024bird}
Yu~Feng, Ben Zhou, Weidong Lin, and Dan Roth. 2024.
\newblock Bird: A trustworthy bayesian inference framework for large language models.
\newblock \emph{arXiv preprint arXiv:2404.12494}.

\bibitem[{Gao et~al.(2023)Gao, Madaan, Zhou, Alon, Liu, Yang, Callan, and Neubig}]{gao2023pal}
Luyu Gao, Aman Madaan, Shuyan Zhou, Uri Alon, Pengfei Liu, Yiming Yang, Jamie Callan, and Graham Neubig. 2023.
\newblock Pal: Program-aided language models.
\newblock In \emph{International Conference on Machine Learning}, pages 10764--10799. PMLR.

\bibitem[{Geva et~al.(2021)Geva, Khashabi, Segal, Khot, Roth, and Berant}]{geva-etal-2021-aristotle}
Mor Geva, Daniel Khashabi, Elad Segal, Tushar Khot, Dan Roth, and Jonathan Berant. 2021.
\newblock \href {https://doi.org/10.1162/tacl_a_00370} {Did aristotle use a laptop? a question answering benchmark with implicit reasoning strategies}.
\newblock \emph{Transactions of the Association for Computational Linguistics}, 9:346--361.

\bibitem[{Hao et~al.(2023)Hao, Gu, Ma, Hong, Wang, Wang, and Hu}]{hao2023reasoning}
Shibo Hao, Yi~Gu, Haodi Ma, Joshua Hong, Zhen Wang, Daisy Wang, and Zhiting Hu. 2023.
\newblock Reasoning with language model is planning with world model.
\newblock In \emph{EMNLP}.

\bibitem[{Hu et~al.(2021)Hu, Wallis, Allen-Zhu, Li, Wang, Wang, Chen et~al.}]{hu2021lora}
Edward~J Hu, Phillip Wallis, Zeyuan Allen-Zhu, Yuanzhi Li, Shean Wang, Lu~Wang, Weizhu Chen, et~al. 2021.
\newblock Lora: Low-rank adaptation of large language models.
\newblock In \emph{ICLR}.

\bibitem[{Hu et~al.(2023)Hu, Storks, Lewis, and Chai}]{Hu2023InContextAR}
Xiaoyang Hu, Shane Storks, Richard~L. Lewis, and Joyce~Yue Chai. 2023.
\newblock In-context analogical reasoning with pre-trained language models.
\newblock In \emph{ACL}.

\bibitem[{Khashabi et~al.(2020)Khashabi, Min, Khot, Sabharwal, Tafjord, Clark, and Hajishirzi}]{khashabi2020unifiedqa}
Daniel Khashabi, Sewon Min, Tushar Khot, Ashish Sabharwal, Oyvind Tafjord, Peter Clark, and Hannaneh Hajishirzi. 2020.
\newblock Unifiedqa: Crossing format boundaries with a single qa system.
\newblock In \emph{ACL (findings)}.

\bibitem[{Khattab et~al.(2022)Khattab, Santhanam, Li, Hall, Liang, Potts, and Zaharia}]{khattab2022demonstrate}
Omar Khattab, Keshav Santhanam, Xiang~Lisa Li, David Hall, Percy Liang, Christopher Potts, and Matei Zaharia. 2022.
\newblock Demonstrate-search-predict: Composing retrieval and language models for knowledge-intensive nlp.
\newblock \emph{arXiv preprint arXiv:2212.14024}.

\bibitem[{Li et~al.(2024)Li, Zhou, Fu, Wang, Roth, and Chen}]{li2024famicom}
Bangzheng Li, Ben Zhou, Xingyu Fu, Fei Wang, Dan Roth, and Muhao Chen. 2024.
\newblock Famicom: Further demystifying prompts for language models with task-agnostic performance estimation.
\newblock \emph{arXiv preprint arXiv:2406.11243}.

\bibitem[{Li et~al.(2023)Li, Zhou, Wang, Fu, Roth, and Chen}]{Li2023Deceiving}
Bangzheng Li, Ben Zhou, Fei Wang, Xingyu Fu, Dan Roth, and Muhao Chen. 2023.
\newblock Deceptive semantic shortcuts on reasoning chains: How far can models go without hallucination?
\newblock \emph{NAACL}.

\bibitem[{Lyu et~al.(2023)Lyu, Havaldar, Stein, Zhang, Rao, Wong, Apidianaki, and Callison-Burch}]{lyu2023faithful}
Qing Lyu, Shreya Havaldar, Adam Stein, Li~Zhang, Delip Rao, Eric Wong, Marianna Apidianaki, and Chris Callison-Burch. 2023.
\newblock Faithful chain-of-thought reasoning.
\newblock In \emph{Proceedings of the 13th International Joint Conference on Natural Language Processing and the 3rd Conference of the Asia-Pacific Chapter of the Association for Computational Linguistics (Volume 1: Long Papers)}, pages 305--329.

\bibitem[{Mirzadeh et~al.(2024)Mirzadeh, Alizadeh, Shahrokhi, Tuzel, Bengio, and Farajtabar}]{mirzadeh2024gsm}
Iman Mirzadeh, Keivan Alizadeh, Hooman Shahrokhi, Oncel Tuzel, Samy Bengio, and Mehrdad Farajtabar. 2024.
\newblock Gsm-symbolic: Understanding the limitations of mathematical reasoning in large language models.
\newblock \emph{arXiv preprint arXiv:2410.05229}.

\bibitem[{OpenAI(2023)}]{Achiam2023GPT4TR}
OpenAI. 2023.
\newblock Gpt-4 technical report.

\bibitem[{Talmor et~al.(2018)Talmor, Herzig, Lourie, and Berant}]{talmor2018commonsenseqa}
Alon Talmor, Jonathan Herzig, Nicholas Lourie, and Jonathan Berant. 2018.
\newblock Commonsenseqa: A question answering challenge targeting commonsense knowledge.
\newblock \emph{arXiv preprint arXiv:1811.00937}.

\bibitem[{Tang et~al.(2023)Tang, Zheng, Li, Meng, Zhu, Liang, and Zhang}]{tang2023large}
Xiaojuan Tang, Zilong Zheng, Jiaqi Li, Fanxu Meng, Song-Chun Zhu, Yitao Liang, and Muhan Zhang. 2023.
\newblock Large language models are in-context semantic reasoners rather than symbolic reasoners.
\newblock \emph{arXiv preprint arXiv:2305.14825}.

\bibitem[{Wei et~al.(2022)Wei, Bosma, Zhao, Guu, Yu, Lester, Du, Dai, and Le}]{weifinetuned}
Jason Wei, Maarten Bosma, Vincent Zhao, Kelvin Guu, Adams~Wei Yu, Brian Lester, Nan Du, Andrew~M Dai, and Quoc~V Le. 2022.
\newblock Finetuned language models are zero-shot learners.
\newblock In \emph{International Conference on Learning Representations}.

\bibitem[{Xu et~al.(2023)Xu, Wang, Zhou, Li, Xiao, and Chen}]{xu2023cognitive}
Nan Xu, Fei Wang, Ben Zhou, Bang~Zheng Li, Chaowei Xiao, and Muhao Chen. 2023.
\newblock Cognitive overload: Jailbreaking large language models with overloaded logical thinking.
\newblock \emph{arXiv preprint arXiv:2311.09827}.

\bibitem[{Yang et~al.(2018)Yang, Qi, Zhang, Bengio, Cohen, Salakhutdinov, and Manning}]{yang2018hotpotqa}
Zhilin Yang, Peng Qi, Saizheng Zhang, Yoshua Bengio, William Cohen, Ruslan Salakhutdinov, and Christopher~D Manning. 2018.
\newblock Hotpotqa: A dataset for diverse, explainable multi-hop question answering.
\newblock In \emph{Proceedings of the 2018 Conference on Empirical Methods in Natural Language Processing}, pages 2369--2380.

\bibitem[{Yao et~al.(2023)Yao, Zhao, Yu, Du, Shafran, Narasimhan, and Cao}]{yaoreact}
Shunyu Yao, Jeffrey Zhao, Dian Yu, Nan Du, Izhak Shafran, Karthik~R Narasimhan, and Yuan Cao. 2023.
\newblock React: Synergizing reasoning and acting in language models.
\newblock In \emph{The Eleventh International Conference on Learning Representations}.

\bibitem[{Yasunaga et~al.(2024)Yasunaga, Chen, Li, Pasupat, Leskovec, Liang, hsin Chi, and Zhou}]{Yasunaga2023LargeLM}
Michihiro Yasunaga, Xinyun Chen, Yujia Li, Panupong Pasupat, Jure Leskovec, Percy Liang, Ed~Huai hsin Chi, and Denny Zhou. 2024.
\newblock Large language models as analogical reasoners.
\newblock \emph{ICLR}.

\bibitem[{Yuan et~al.(2024)Yuan, Chen, Sun, Liang, Xiao, and Yang}]{Yuan2023ANALOGYKBUA}
Siyu Yuan, Jiangjie Chen, Changzhi Sun, Jiaqing Liang, Yanghua Xiao, and Deqing Yang. 2024.
\newblock Analogykb: Unlocking analogical reasoning of language models with a million-scale knowledge base.
\newblock \emph{ACL}.

\bibitem[{Zelikman et~al.(2024)Zelikman, Harik, Shao, Jayasiri, Haber, and Goodman}]{zelikman2024quiet}
Eric Zelikman, Georges Harik, Yijia Shao, Varuna Jayasiri, Nick Haber, and Noah~D Goodman. 2024.
\newblock Quiet-star: Language models can teach themselves to think before speaking.
\newblock \emph{arXiv preprint arXiv:2403.09629}.

\bibitem[{Zhou et~al.(2024)Zhou, Zhang, Chen, Yu, Wang, Peng, Roth, and Yu}]{Zhou2023Symbolic}
Ben Zhou, Hongming Zhang, Sihao Chen, Dian Yu, Hongwei Wang, Baolin Peng, Dan Roth, and Dong Yu. 2024.
\newblock Conceptual and unbiased reasoning in language models.
\newblock \emph{arXiv preprint arXiv:2404.00205}.

\bibitem[{Zhu et~al.(2024)Zhu, Qi, Zhang, Long, Lin, and Zhou}]{zhu2024pad}
Xuekai Zhu, Biqing Qi, Kaiyan Zhang, Xinwei Long, Zhouhan Lin, and Bowen Zhou. 2024.
\newblock Pad: Program-aided distillation can teach small models reasoning better than chain-of-thought fine-tuning.
\newblock In \emph{Proceedings of the 2024 Conference of the North American Chapter of the Association for Computational Linguistics: Human Language Technologies (Volume 1: Long Papers)}, pages 2571--2597.

\end{thebibliography}
\bibliographystyle{acl_natbib}

\clearpage
\appendix

\section{Prompts}
\label{sec:prompts}

\subsection{Few-shot Prompts for Seed Supervision}
Table~\ref{tab:p1} lists the prompts we use for collecting seed supervision of math question abstractions. Non-math question abstractions are done in \citet{Zhou2023Symbolic}, and we directly borrow their data. Table~\ref{tab:p2} shows the prompts we use to collect seed supervision for generating gold math questions. The generated programs' validity is further verified by the gold answer. Table~\ref{tab:p3} details the prompt for collecting seed supervision data for generating the next questions in the simplification pipeline. This prompt asks for the entire decomposition with multiple next questions in one query, later formulated as multiple instances with one next question per instance. Table~\ref{tab:p4} shows the prompts for collecting seed supervision instances on known conditions. It handles two scenarios: extracting the list of known conditions from the original question and transforming the answer to the next question into a statement. These two scenarios are formulated as separate seed supervision instances.

\begin{table*}[ht]
\centering
\small
\begin{tabular}{p{0.08\linewidth} | p{0.85\linewidth}}
\toprule
Prompt & Content \\
\cmidrule(lr){1-1}\cmidrule(lr){2-2}
system & Identify numerical values in the given question, then replace some of them with Python parameters that are either int or float, so that the resulting abstract question is still answerable with the same general solution as the original question. Follow the provided examples. \\
user & Jack has a stack of books that is 12 inches thick. He knows from experience that 80 pages is one inch thick. If he has 6 books, how many pages is each one on average? \\
assistant & Becuase this is a math question, we identify all numerical values. We identify: 12 inches, 80 pages, one inch, 6 As a result, we can replace "12 inches" to "Number of Inches X" (num$\_$inches$\_$x: int) "80 pages" to "Number of Pages Y" (num$\_$pages$\_$y: int) "one inch" to "Number of Inches Z" (num$\_$inches$\_$z: int) "6" to "Number W" (num$\_$w: int) So the question becomes Jack has a stack of books that is Number of Inches X thick. He knows from experience that Number of Pages Y is Number of Inches Z thick. If he has Number W books, how many pages is each one on average? With parameters num$\_$inches$\_$x=12, num$\_$pages$\_$y=80, num$\_$inches$\_$z=1, num$\_$w=6 \\
\bottomrule
\end{tabular}
\caption{Prompts for collecting seed supervision of math question abstractions. More examples are omitted.}
\label{tab:p1}
\end{table*}

\begin{table*}[ht]
\centering
\small
\begin{tabular}{p{0.08\linewidth} | p{0.85\linewidth}}
\toprule
Prompt & Content \\
\cmidrule(lr){1-1}\cmidrule(lr){2-2}
system & Based on the given question, write a Python program with some abstraction that solves the given question and all other similar questions that can be solved in a similar fashion. Think as comprehensively as possible, so that the program would work on any inputs. You can use the following helper function: ask$\_$gpt(query: str, type) to ask a simple question to the knowledgeable GPT model and returns the answer in the desired type. \\
user & Benny bought Number of Soft Drinks X (num$\_$soft$\_$drinks$\_$x: int) for Cost per Soft Drink Y (cost$\_$per$\_$soft$\_$drink$\_$y: int) each and Number of Candy Bars Z (num$\_$candy$\_$bars$\_$z: int). He spent a total of Total Amount Spent W (total$\_$spent$\_$w: int) dollars. How much did each candy bar cost?  \\
assistant & \begin{python}
def answer(num_soft_drinks_x: int, cost_per_soft_drink_y: int, num_candy_bars_z: int, total_spent_w: int) -> float:
	total_cost_soft_drinks = num_soft_drinks_x * cost_per_soft_drink_y
	total_cost_candy_bars = total_spent_w - total_cost_soft_drinks
	cost_candy_bar = total_cost_candy_bars / num_candy_bars_z
	return cost_candy_bar
\end{python} \\

\bottomrule
\end{tabular}
\caption{Prompts for collecting seed supervision of math programs. More examples are omitted.}
\label{tab:p2}
\end{table*}

\begin{table*}[ht]
\centering
\small
\begin{tabular}{p{0.08\linewidth} | p{0.85\linewidth}}
\toprule
Prompt & Content \\
\cmidrule(lr){1-1}\cmidrule(lr){2-2}
system & Given a math question, your job is to decompose it into sub-questions that can stack together and lead to the correct answer of the original question. Each sub question should be answerable on its own. Use '\#\#\#\#' to mark your actual answer when answering each sub question. Reference the correct answer. \\
user & Benny bought 2 soft drinks for\$ 4 each and 5 candy bars. He spent a total of 28 dollars. How much did each candy bar cost? \\
assistant & Question 1: Benny bought 2 soft drinks for\$ 4 each. How much does the soft drinks cost in total? Answer 1: The total cost of the soft drinks are <<2*4=8>> \#\#\#\# 8 Question 2: Benny spent \$8 on soft drinks and bought 5 candy bars. He spent a total of 28 dollars. How much did the candy bars cost? Answer 2: The candy bars cost <<28-8=20>> \#\#\#\# 20 Question 3: Benny spent \$20 to buy 5 candy bars. How much did each candy bar cost? Answer 3: Each of the candy bars cost <<20/5=4>> \#\#\#\# 4 END \\
user & There are some lions in Londolozi at first. Lion cubs are born at the rate of 5 per month and lions die at the rate of 1 per month. If there are 148 lions in Londolozi after 1 year, how many lions were there in Londolozi at first? \\
assistant & Question 1: There are some lions in Londolozi at first. Lion cubs are born at the rate of 5 per month and lions die at the rate of 1 per month. What is the number of new lions per months? Answer 1: There are <<5-1=4>> new lions per months. \#\#\#\# 4 Question 2: There are some lions in Londolozi at first. There are 4 new lions every months. If there are 148 lions in Londolozi after 1 year, how many lions were there in Londolozi at first? Answer 2: The number of new lions in one year is <<4*12=48>>. The original lions are <<148-48=100>> \#\#\#\# 100 END \\
\bottomrule
\end{tabular}
\caption{Prompts for collecting seed supervision of math next questions. More examples are omitted.}
\label{tab:p3}
\end{table*}

\begin{table*}[ht]
\centering
\small
\begin{tabular}{p{0.08\linewidth} | p{0.85\linewidth}}
\toprule
Prompt & Content \\
\cmidrule(lr){1-1}\cmidrule(lr){2-2}
system & Given a math question, extract all the known conditions that are explicitly mentioned. If the answer is provided, extract only the condition that the answer implies. \\
user & Question: Benny bought  2 soft drinks for\$ 4 each and 5 candy bars. He spent a total of 28 dollars. How much did each candy bar cost? Answer: unknown \\
assistant & Benny bought 2 soft drinks for \$4 each. Benny bought 5 candy bars. Benny spent 28 dollars in total. \\
user & Question: Wickham is throwing a huge Christmas party. He invites 30 people. Everyone attends the party, and half of the guests bring a plus one (one other person). He plans to serve a 3-course meal for the guests. If he uses a new plate for every course, how many plates does he need in total for his guests? Answer: unknown \\
assistant & Wickham invites 30 people. Half of the guests bring a plus one. Wickham prepares 3 courses for each person. He uses a new plate for each course. \\
user & Question: If lion cubs are born at the rate of 5 per month and lions die at the rate of 1 per month, what is the net increase in the number of lions per month? Answer: 4 \\
assistant & The number of lines increase by 4 every month. \\
user & Question: It takes 7 years for an apple tree to bear fruit. If Lydia planted a tree when she was 4 years old, how old would she be when the tree bears fruit for the first time? Answer: 11 \\
assistant & Lydia would be 11 years old when the tree bears fruit for the first time. \\
user & Question: Wickham invites 30 people to the party. Half of the guests bring a plus one. How many additional people (plus ones) are coming to the party? Answer: 15 \\
assistant & 15 additional people are coming to the party. \\
\bottomrule
\end{tabular}
\caption{Prompts for collecting seed supervision of extracting initial known conditions and transforming a question and its answer to a statement. ``unknown'' answers refer to the initial known condition generation scenarios.}
\label{tab:p4}
\end{table*}

\subsection{Input Templates}

Table~\ref{tab:t1} details the input prompt templates for each sub-task as described in \S\ref{sec:overview-conceptualization} and \S\ref{sec:overview-simplification}. A multi-function model trained with the same prompt templates can recognize the different goals based on the query because the seed supervision instances use the same templates.

\begin{table*}[ht]
\centering
\small
\begin{tabular}{p{0.16\linewidth} | p{0.74\linewidth}}
\toprule
Subtask & Content \\
\cmidrule(lr){1-1}\cmidrule(lr){2-2}
$\mathrm{cot}(q)$ & Answer this question by thinking step by step. Question: \{q\} Output: \\
$\mathrm{abs}(q)$ & Convert this question to the abstract form. Question: \{q\} Output: \\
$\mathrm{q2p}(q)$ & Solve this question with a Python program with necessary abstractions. Also generate the corresponding values for the parameters. Question: \{q\} Output: \\
$\mathrm{aq2p}(aq)$ & Solve this abstract question with a Python program. Abstract Question: \{aq\} Output: \\ 
$\mathrm{qc2p}(q, cot)$ & Solve this question with a Python program based on the provided reference answer. Question: \{q\} Reference Answer: \{cot\} Output: \\
$\mathrm{math\_q2kc}(q)$ & Extract the given conditions in the math question. Question: \{q\} Output: \\
$\mathrm{math\_q2nq}(q, c)$ & Decompose the math question to steps, based on currently known conditions. Question: \{q\} Known Conditions: \{c\} Output: \\
$\mathrm{qa2s}(q, a)$ & Convert the given math question and its answer to a statement. Question: \{q\} Answer: \{a\} Output: \\
\bottomrule
\end{tabular}
\caption{Input prompt templates for each sub-task that the multi-function model handles.}
\label{tab:t1}
\end{table*}

\section{Implementations}
\label{sec:implementations}
Table~\ref{tab:t2} shows a partial implementation of our programmatic solution space. In addition to this, we also employ the soft value comparison helper functions as described in \citet{Zhou2023Symbolic}.

All SFT experiments are conducted with 4x RTX6000 GPUs and a single-run fine-tuning finishes in 3 hours. We useHuggingface's Autotrain framework for training.\footnote{\url{https://github.com/huggingface/autotrain-advanced}} All parameters are default parameters unless explicitly mentioned otherwise.

\begin{table*}[ht]
\centering
\small
\begin{tabular}{p{0.85\linewidth}}
\toprule
\begin{python}
def query_llm(messages):
	# library function that makes a query to an LLM.

def get_unit(question):
    messages = [
        {"role": "user", "content": "Does the answer to the following question require a measurement unit? If it is physical metric, use SI units. If it is monetary metric, use 'USD'.\nQuestion: What is the density of copper?\nAnswer with the proper unit, or 'None' if not applicable."},
        {"role": "assistant", "content": "kg/m^3"},
        {"role": "user", "content": "More examples omitted"},
        {"role": "user", "content": "Can the following question be answered with a metric in the International System of Units (SI Units)?\nQuestion: {}\nAnswer with the proper unit, or 'None' if not applicable.".format(question)},
    ]
    result = query_llm(messages)
    if 'None' not in result:
        return result
    return None

def find_json(s):
    # finds valid JSON structure in the output

def ask_llm(question, t, use_cache=False):
    unit = get_unit(question)
   	question += " Answer in {}.".format(unit)
    messages = [
        {"role": "system", "content": "Answer the question in the expected type. Use your best educated guess or estimation if needed."},
        {"role": "user", "content": "How many people today are related to Genghis Khan? (int)"},
        {"role": "assistant", "content": "{\"answer\": 35000000}"},
        {"role": "user", "content": "More examples omitted"},
        {"role": "user", "content": "{} ({})".format(question, t.__name__)}
    ]
    response = query_llm(messages)
    r = find_json(response)
    return r
\end{python}
\end{tabular}
\caption{Implementations of the helper functions in our programmatic solution space. Some details are omitted for simplicity.}
\label{tab:t2}
\end{table*}

\end{document}